\documentclass[runningheads]{llncs}
\usepackage{graphicx}
\usepackage{amsmath,amssymb} 
\usepackage{color}
\usepackage[width=122mm,left=12mm,paperwidth=146mm,height=193mm,top=12mm,paperheight=217mm]{geometry}
\usepackage{epstopdf}
\usepackage{multirow}

\graphicspath{{./images/}}
\newcommand{\argmax}{\operatornamewithlimits{argmax}}

\def\bh{{\bf h}}

\def\bp{{\bf p}}

\def\bq{{\bf q}}

\def\bbf{{\bf f}}
\def\bx{{\bf x}}

\def\bY{{\bf Y}}

\begin{document}
\pagestyle{headings}
\mainmatter

\title{Learning Dynamic Hierarchical Models for Anytime Scene Labeling} 

\titlerunning{Learning Dynamic Hierarchical Models for Anytime Scene Labeling}

\authorrunning{Buyu Liu and Xuming He}

\author{Buyu Liu$^{1,2}$ and Xuming He$^{2,1}$}


\institute{$^1$The Australian National University, $^2$Data61, CSIRO\\
	\email{ \{buyu.liu,xuming.he\}@anu.edu.au}
}

\maketitle

\begin{abstract}
With increasing demand for efficient image and video analysis, test-time cost of scene parsing becomes critical for many large-scale or time-sensitive vision applications.
We propose a dynamic hierarchical model for anytime scene labeling that allows us to achieve flexible trade-offs between efficiency and accuracy in pixel-level prediction. 
In particular, our approach incorporates the cost of feature computation and model inference, and optimizes the model performance for any given test-time budget by learning a sequence of image-adaptive hierarchical models.
We formulate this anytime representation learning as a Markov Decision Process with a discrete-continuous state-action space. A high-quality policy of feature and model selection is learned based on an approximate policy iteration method with action proposal mechanism. We demonstrate the advantages of our dynamic non-myopic anytime scene parsing on three semantic segmentation datasets, which achieves $90\%$ of the state-of-the-art performances by using $15\%$ of their overall costs.

\end{abstract}
\section{Introduction}

A fundamental and intriguing property of human scene understanding is its efficiency and flexibility, in which vision systems are capable of interpreting a scene at multiple levels of details given different time budgets~\cite{hegde2008time,Karayev2014}. Despite much progress in the pixel-level semantic scene parsing~\cite{gould2014scene,long2014fully,hariharan2014hypercolumns,dai2014convolutional}, most efforts are focused on improving the prediction accuracy with complex structured models~\cite{he2004multiscale,YaoFU12} and learned representations~\cite{farabet2013learning,chen2014semantic,simonyan2014very}. Such computation-intensive approaches often lack the flexibility in trade-off between efficiency and accuracy, making it challenging to apply them to large-scale data analysis or cost-sensitive applications. 


In order to improve the efficiency in scene labeling, a common strategy is to develop active inference mechanisms for the structured models used in this task~\cite{Roig2013,Weiss2013}. This allows users to adjust the trade-off between efficiency and accuracy for a \textit{given} model, which is learned using a separate procedure with unconstrained test-time budget. However, this may lead to a sub-optimal performance for the cost-sensitive tasks. 

A more appealing approach is to learn a model representation \textit{for} Anytime performance, which can stop its inference at any cost budget and achieve an optimal prediction performance under the cost constraint~\cite{zilberstein1996using}. While such learned representations have shown promising performance in anytime prediction, most work address the unstructured classification problems and focus on efficient feature computation~\cite{Karayev2014,xu2013anytime,grubb2012speedboost}. Only recent work of Grubb et al.~\cite{Grubb2013} proposes an anytime prediction method for scene parsing, which relies on learning a representation 
for individual segments. Nevertheless, to achieve coherent scene labeling, it is important to learn a representation that also encodes the relations between scene elements (e.g., segments).  


\begin{figure}[t!]
	\centering
	\includegraphics[width=0.8\textwidth]{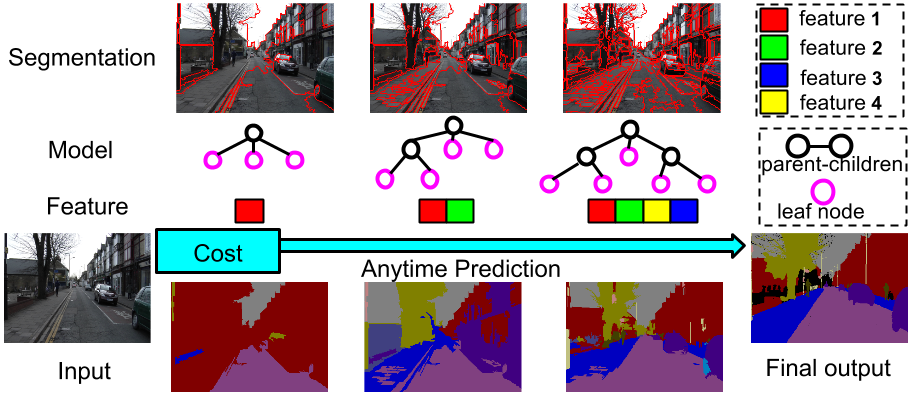}
	\caption{\small {Overview of our approach. {We propose to incrementally increase model complexity in terms of used image features and model structure. Our approach generates high-quality prediction at any given cost.}}}\label{fig:overview}
\end{figure}


In this work, we tackle the anytime scene labeling problem by learning a family of structured models that captures long-range dependency between image segments. Labeling with structured models, however, involves both feature computation \textit{and} inference cost. To enable anytime prediction, we propose to generate scene parsing from spatially coarse to fine level and with increasing number of image features. Such a strategy allows us to control both feature computation cost \textit{and} the model structure which determines the inference cost. 

Specifically, we design a hierarchical model generation process based on growing a segmentation tree for each image. Starting from the root node, this process gradually increases the overall model complexity  by either splitting a subset of leaf-node segments or adding new features to label predictors defined on the leaf nodes. At each step, the resulting model encodes the structural dependency of labels in the hierarchy. 
For any cost budget, we can stop the generation process and produce a scene labeling by collecting the predictions from leaf nodes.    
An overview of our coarse-to-fine scene parsing is shown in Figure~\ref{fig:overview}. We note that a large variety of hierarchical models can be generated with different choices of node splitting and feature orders.

To achieve aforementioned Anytime performance, we seek a policy of generating the hierarchical models which produce high-quality or optimal pixel-level label predictions for any given test budget. We follow the anytime setting in~\cite{grubb2012speedboost,xu2013anytime}, in which the test-time budget is \textit{unknown} during model learning. 
Instead of learning a greedy strategy, we formulate the anytime scene labeling as a sequential decision making problem, and define a finite-horizon Markov Decision Process (MDP). The MDP  maximizes the average label accuracy improvement per unit cost (or the average `speed' of improvement if the cost is time) over a range of cost budgets as a surrogate for the anytime objective, and has a parametrized discrete action space for expanding the hierarchical models. 

We solve the MDP to obtain a high-quality policy by developing an approximate least square policy iteration algorithm~\cite{lagoudakis2003least}.  
To cope with the parametrized action space, we propose an action proposal mechanism to sample a pool of candidate actions, of which the parameters are learned based on several greedy objectives and on different subsets of images. 
We note that the key properties of our learned policy are \textit{dynamic}, which generates an image-dependent hierarchical representation, and \textit{non-myopic}, which takes into account the potential future benefits in a sequence of predictions.    

We evaluate our dynamic anytime parsing method on three publicly available semantic segmentation benchmarks, CamVid~\cite{BrostowSFC:ECCV08}, Stanford Background~\cite{Gould:ICCV09} and Siftflow~\cite{liu2011nonparametric}. The results show that our approach is favorable in terms of anytime scene parsing compared to several state-of-the-art representation learning strategies, and in particular we can achieve $90\%$ of the state-of-the-art performances within $15\%$ of their total costs.

\section{Related work}
Semantic scene labeling has become a core problem in computer vision research~\cite{gould2014scene}. While early efforts tend to focus on structural models with hand-crafted features, recent work shift towards deep convolutional neural network based representation with significant improvement on prediction accuracy~\cite{long2014fully,chen2014semantic,dai2014convolutional}. Hierarchical models, such as dynamic trees~\cite{slorkey2003image}, segmentation hierarchies~\cite{socher2011parsing,lempitsky2011pylon,russell2009associative,munoz2010stacked} and And-Or graphs~\cite{zhu2007stochastic}, have adopted for semantic parsing. However, in general, those methods are expensive to deploy due to complex model inference or costly features. 

Most of prior work on efficient semantic parsing focus on the active inference, which assumes redundancy in pre-learned models and achieves efficiency by allocating resource to an informative subset of model components. Roig et al.~\cite{Roig2013} use perturb-and-MAP inference model to select informative unary potentials to compute. 
Liu and He~\cite{Liu:CVPR2015} actively select most-rewarding subgraphs for video segmentation. In~\cite{Riemenschneider2014}, a local classifier is learned to select views for multi-view semantic labeling. 
Unlike these methods, we explicitly learn a representation for achieving strong performance at any test-time budget. 

Learning anytime representation has been extensively explored for unstructured prediction problems (e.g., classification)~\cite{xu2013anytime,grubb2012speedboost}. 
Karayev et al.~\cite{Karayev2014} learn an anytime representation for object and scene recognition, focusing on dynamic feature selection under a total budget.  
Weiss and Taskar~\cite{Weiss2013} develop a reinforcement learning framework for feature selection in structured models. 
In contrast, we consider both feature computation and model inference cost. More importantly, we incorporate the cost in an MDP reward which encourages anytime property. Unlike~\cite{Karayev2014}, the test-time budget is explicitly unknown during learning in our setting. 
Perhaps the most related work is~\cite{Grubb2013}, which learns a segment-based anytime representation consisting of a selection function and a boosted predictor for individual segments. Their policy of segment and feature selection is trained in a greedy manner based on~\cite{grubb2012speedboost} and a single strategy is applied to all the images. By contrast, we build a structured hierarchical model on segmentation trees and learn an image-adaptive policy.

More generally, cost-sensitive learning and inference have been widely studied in learning and vision literature under various different contexts, including feature selection~\cite{he2013dynamic}, learning classifier cascade by empirical risk minimization~\cite{wang2014model,trapeznikov2013supervised} or Wald's sequential ratio test~\cite{sochman2005waldboost}, model selection~\cite{Weiss2013b,benbouzid2012fast}, prioritized message passing inference~\cite{jiangprioritized}, object detection~\cite{DecisionPolicy_ICCV2013}, and activity recognition~\cite{Amer2012}. However, few approaches have been designed for optimizing the anytime prediction performance~\cite{zilberstein1996using}, or considering both feature and inference costs. 
We note that while the MDP framework has been extensively used in those methods, our formulation of discrete-continuous MDP is tailored for anytime scene parsing. 


Unfolding and learning inference in graphical models has been explored in various inference machines~\cite{munoz2010stacked,ross2011learning}. Nevertheless, such methods usually use a greedy approach to learn the messages or model predictions. \cite{denoyer2014deep} use reinforcement learning to obtain a dynamic deep network model, but they do not address the structured prediction problem. 
Lastly, we note that, although some search-based structured prediction methods~\cite{doppa2014structured,zhang2014greed} are capable of terminating inference and generating outputs at any time, they usually do not consider feature computation cost and are not optimized for anytime performance.


\setlength{\belowdisplayskip}{1.5mm} 
\setlength{\abovedisplayskip}{1.5mm}
\section{Anytime scene labeling with a hierarchical model}\label{sec:model}

We aim to learn a structured model representation with anytime performance property for semantic scene labeling. As structured prediction involves both feature computation and inference, we need a flexible representation that allows us to control the cost of feature and inference computation. To this end, we first introduce a family of hierarchical models based on image segmentation trees in Sec~\ref{subsec:hierarchy}, which is capable of incrementally increasing its complexity in terms of used image features and model structure.  

We then formulate the anytime scene labeling as a sequential feature and model selection process in this model family with a cost-sensitive labeling loss in Sec~\ref{subsec:anytime} and Sec~\ref{subsec:reward}. Based on an MDP framework, our goal is to learn an optimal selection policy to generate a sequence of hierarchical models from a set of annotated images. In Sec~\ref{sec:learn}, we develop an iterative procedure to solve the policy learning problem approximately.   


\subsection{Coarse-to-fine scene parsing with a segmentation hierarchy}\label{subsec:hierarchy}

We now introduce a flexible hierarchical representation for semantic parsing that enables us to control the test-time complexity. 
To achieve effective semantic labeling, we want to design a model framework capable of incorporating rich image features, modeling long-range dependency between regions and achieving anytime property.
To this end, we adopt a coarse-to-fine scene labeling strategy, and consider a family of hierarchical models built on image segmentation trees, which has a simplified form of the Hierarchical Inference Machine (HIM)~\cite{munoz2010stacked}.  


Specifically, given an image $I$, we construct a sequence of segmentation trees by recursively partitioning the image using graph-based algorithms~\cite{felzenszwalb2004efficient,grundmann2010efficient}. 
We then develop a sequence of hierarchical models that predict label marginal distributions on the leaf nodes of the segmentation trees. 
Formally, let the semantic label space be $\mathcal{Y}$. We start from an initial segmentation tree  $\mathcal{T}^0$ with a single node and a marginal distribution $\mathcal{Q}^0=\{\bq^0\}$ on the node, which can be uniform or a global label prior. 
We incrementally grow the tree and update the prediction of marginal distributions on the leaf nodes by two update operators described in detail below, which generates a sequence of hierarchical models for labeling, denoted by $\mathcal{M}^1, \cdots, \mathcal{M}^T$, where $T$ is the total number of steps.  

At each step $t$, the hierarchical model $\mathcal{M}^t$ consists of a tree $\mathcal{T}^t$ and a set of predicted label distributions on the tree's leaf nodes, $\mathcal{Q}^t$. More concretely, we denote the leaf nodes of $\mathcal{T}^t$ as $\mathcal{B}^t=\{b_{1},\cdots,b_{N_t}\}$ where $N_t$ is the number of leaf nodes. We associate each leaf-node segment $b_{i}$ with a label variable $y_i^t$ indicating its dominant label assignment. Let the label distributions $\mathcal{Q}^{t} = \{\bq_i^{t}\}_{i=1}^{N_t}$, where $\bq_i^t$ is the current label marginals at node $b_i$. We generate the next hierarchical model $\mathcal{M}^{t+1}$ by applying the following two update operators. 




\noindent\textbf{Split-inherit update.} We choose a subset of leaf-node segments and split them into finer scale segments in the segmentation tree. The selection criterion is based on the entropy of the node marginals $H(\bq_i^{t})$, and all the nodes with $H(\bq_i^{t})>\theta_t$ will split into their children~\cite{Grubb2013}. $\theta_t$ is a parameter of the operator and $\theta_t\in\mathbb R$. 
The new leaf-node segments inherit the marginal distributions of their parents. 
\begin{equation}\label{eqn:split}
\bq_i^{t+1}(k) = \bq_{pa(i)}^{t}(k),\quad k\in\mathcal{Y}, \; i\in \mathcal{B}^{t+1}
\end{equation}
where $\mathcal{B}^{t+1}$ is the new leaf node set and $pa(i)$ indicates the parent node of $i$ in the new tree $\mathcal{T}^{t+1}$. We denote the parameter space of the operator as $\varTheta$. 

\noindent\textbf{Local belief update.} For the newly generated leaf nodes from splitting, we improve their marginal distributions by 
adding more image cues or context information from their parents. Specifically, we extract a set of input features $\bx_i$ from segment $b_i$, and adopt a boosting-like strategy: Using a weak learner 
taking the image feature $\bx_i$ and the marginal of its parent $\bq_{pa(i)}^{t}$ as input~\cite{munoz2010stacked}, we update the marginals of leaf nodes as follows, 
\begin{equation}\label{eqn:local}
\bq_i^{t+1}(k) \propto {\bq}_{i}^{t}(k)\exp\left(\alpha_t h_k^t(\bbf_{i}^t(j))\right), \quad k\in\mathcal{Y}, \quad \bbf_i^t=[\bx_i, \bq_{pa(i)}^{t}]
\end{equation}
where $\bbf_{i}^t(j)$ is the $j$-th feature used in the weak learner; $\bh_t=[h_t^1,...,h_t^{|\mathcal{Y}|}]$ and $\alpha_t$ are the newly added weak learners and their coefficient, respectively. We denote the weak learner space as $\mathcal{H}$ and $\alpha_t\bh_t\in\mathcal{H}$.

\begin{figure}[t!]
	\centering
	\includegraphics[width=0.75\textwidth]{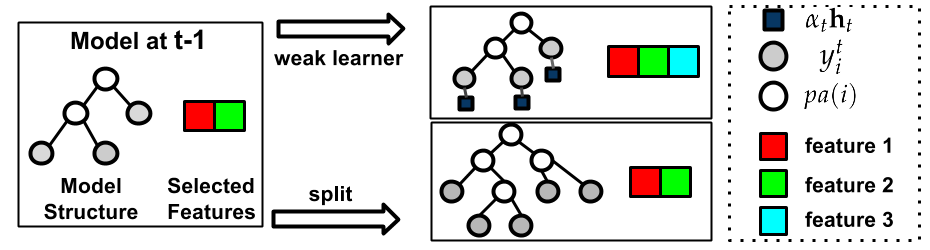}
	\caption{\small {{Example of operators at step $t$. We choose either to add weak learners or split a subset of leaf nodes, which leads to gradually increasing model complexity.}}}\label{fig:model_overview}
\end{figure}

By applying a sequence of these update operators to the segmentation tree from its root node, we can generate a dynamically growing hierarchical models for scene labeling (see Fig~\ref{fig:model_overview} for an illustration). We refer to the resulting structured models as the \textit{Dynamic Hierarchical Model} (DHM). 
We use `dynamic' to indicate that our model generation process can vary from image to image or given different choices of the operators, which is not predetermined by greedy learning as in~\cite{munoz2010stacked}. 
Using DHM as our representation for anytime scene labeling has several advantages. First, a DHM is capable of generating a sequence of model predictions with incrementally increasing cost as every update operator can be computed efficiently. In addition, it utilizes multiscale region grouping to create models from coarse to fine level, leading to gradually increasing model complexity. Furthermore, it has a flexible structure to select image features by weak learners and to capture long-range dependency between segments, which is critical to achieve the state-of-the-art performance for any test-time budget. 
\subsection{Anytime scene labeling by cost-sensitive DHM generation}\label{subsec:anytime}

Given the dynamic hierarchical scene models defined in Sec~\ref{subsec:hierarchy}, we now formulate the anytime scene labeling as a cost-sensitive DHM generation problem. Specifically, 
we want to find a model generation strategy, which selects an sequence of image-dependent update operators, such that the incrementally built hierarchical models achieve good performance (measured by average labeling accuracy) at all possible test-time cost budgets.    
To address this sequential selection problem, we model the cost-sensitive model generation as a Markov Decision Process (MDP) that encourages good anytime performance with a cost-sensitive reward function. By solving this MDP, we are able to find a policy of selection that yields a sequence of hierarchical models with high-quality anytime performance.

Concretely, we first model an episode of coarse-to-fine DHM generation as an MDP with finite horizon. This MDP consists of a tuple $\{\mathcal{S}, \mathcal{A},T(\cdot), R(\cdot),\allowbreak \gamma\}$, which defines its state space, action space, state transition, reward function and a discounting factor, respectively.

\textbf{State:} At time $t$, the state $s_t\in\mathcal{S}$ represents the current segment set corresponding to the leaf nodes of the segmentation tree and the label marginal distributions on the leaf nodes. As in Sec~\ref{subsec:hierarchy}, we denote the leaf-node segment set and the corresponding marginal label distributions as $\mathcal{B}^t=\{b_i\}^{N_t}_{i=1}$ and $\mathcal{Q}^t=\{\bq^t_i\}^{N_t}_{i=1}$ respectively. We also introduce an indicator vector $Z^t\in\{0,1\}^{N_t}$ to describe an active set of leaf-nodes at $t$, in which $Z^t(k)=1$ indicates the leaf node $b_k$ is newly generated by splitting. Altogether, we define $s_t=\{\mathcal{B}^t,\mathcal{Q}^t,Z^t\}$.

\textbf{Action:} The action set $\mathcal{A}$ consists of the two types of update operators defined in Sec~\ref{subsec:hierarchy}. We denoted them by $\{u_s(\theta),u_b(\alpha\bh)\}$. For $a_t=u_s(\theta_t)$, we choose to split a subset of leaf-node segments of which the entropy of predicted marginal distributions are greater than $\theta_t$. For $a_t=\alpha_t\bh_t$, we apply the local belief update to the active nodes in $Z^t$ using the weak learner $\alpha_t\bh_t\in \mathcal{H}$. 
Note that the action space $\mathcal{A}$ is a discrete-continuous space $\varTheta\cup\mathcal{H}$ due to their parameterization.

\textbf{State Transition:} The state transition $T(s_{t+1}|s_t,a_t)$ is a deterministic function in our MDP. For $a_t=\theta_t$, it expands the tree and generates a new set of leaf-node segments $\mathcal{B}^{t+1}$ with inherited marginals $\mathcal{Q}^{t+1}$ as defined in Eqn~\eqref{eqn:split}. The new active regions are the newly generated leaf-nodes from splitting, denoted by $Z^{t+1}$. The action $a_t=\alpha_t\bh_t$ keeps the tree structure and active regions unchanged, such that $\mathcal{B}^{t+1}=\mathcal{B}^{t}$ and $Z^{t+1}=Z^t$; while it only updates the node marginals $\mathcal{Q}^{t+1}$ according to Eqn~\eqref{eqn:local}. 

\textbf{Reward Function and $\gamma$:} The reward function $R$ defines a mapping from $(s_t,a_t)$ to rewards in $\mathbb{R}$ and {$\gamma$ is a discount factor that determines the lookahead in selection actions. For the  anytime learning problem, we design a reward function that is cost-sensitive and encourages the sequence of generated models can achieve good labeling accuracies across a range of possible cost budgets. The details of the reward function and $\gamma$ will be discussed in the next subsection. 


\subsection{Defining reward function}~\label{subsec:reward}

We now define a reward function that favors a coarse-to-fine dynamic hierarchical model generation with anytime performance. To this end, we first describe the action costs of the MDP, which compute the overall cost of model prediction. We then introduce a labeling loss for our hierarchical models, based on which the cost-sensitive reward and the value function of the MDP are defined. 

\textbf{Action Cost:} The action cost represents the cost of scene labeling using a hierarchical model, which consists of feature extraction cost $c_{f_t}$ for computing feature set $f_t$ (from the entire image or specific regions), region split cost $c_r$ for pooling features for newly split regions, total weak learner cost $c_{h_t}$ for applying the weak learner $\alpha_t\bh_t$ to predict labels. For each action $a_t$, we define the action cost $c(a_t)$ as $c_{h_t}+c_{f_t}$ if $a_t = u_b$, or $c_r$ if $a_t = u_s$. In this work, we use the CPU time used in $a_t$ as a surrogate for the computation cost while any other type of costs can also be applied. 

\textbf{Labeling loss of DHMs:} Given a hierarchical model represented by $s_t$, we introduce a loss function measuring the scene labeling performance. Particularly, we adopted an entropy-based labeling loss function defined as follows, 
	\begin{equation}\label{eqn:energy}
	\mathcal{L}(s_{t},\hat{\bY}|I)=-\sum_{i\in \mathcal{B}^t} w_i \bp_i^T \log(\bq^t_i)-\alpha \sum_{i\in \mathcal{B}^{t}} w_i \bp_i^T\log(\bp_i), 
	\end{equation}
where $\bp_i$ is the ground-truth label distribution in region $b_i\in\mathcal{B}^t$, derived from the ground-truth labeling $\hat{\bY}$, and $w_i$ denotes its normalized size. The first term is the cross-entropy between the marginals and the ground-truth while the second term penalizes the regions with mixed labels. These two terms reflect the label prediction and image partition quality respectively and we further introduce a weight $\alpha$ to control their balance. Intuitively, the loss favors a model with a sensible image segmentation and a good prediction for the segment labels. A larger $\alpha$ prefers to learn predictors after reaching fine levels in hierarchy while a small value may lead to stronger predictors at all levels. 

\textbf{Cost-sensitive reward:}	To achieve the anytime performance, an ideal model generation sequence will minimize the labeling loss as fast as possible such that it can obtain high quality scene labeling for a full range of cost budgets. Following this intuition, we define the reward for action $a_t$ as the labeling loss improvement between $s_{t+1}$ and $s_t$ normalized by the cost of $a_t$~\cite{grubb2012speedboost}. Formally, we define the reward as,
\begin{equation}\label{eq:reward}
R(s_t,a_t|I)=\frac{1}{c(a_t)}\left[\mathcal{L}(s_t,\hat{\bY}|I)-\mathcal{L}(s_{t+1},\hat{\bY}|I)\right]
\end{equation}
where $c(a_t)$ summarizes all the computation cost in the action $a_t$. 

\textbf{Policy and value function:} A policy of the MDP is a function mapping from a state to an action, $\pi(s):\mathcal{S}\rightarrow\mathcal{A}$. The value function of the MDP at state $s_t$ under policy $\pi$ is the total accumulated reward defined as,
\begin{align}\label{eq:value}
V_{\pi}(s_t)=\sum_{\tau=t}^T\gamma^{\tau-t}R(s_\tau,\pi(s_\tau)|I)
\end{align}
where $T$ is the number of actions taken and $s_0$ is the initial state. 
Our goal is to find an optimal policy $\pi^*$ that maximizes the expected value function over the image space for any state $s_t$. We will discuss how to learn such a policy using a training set in the following section. We note that our objective describes a weighted average speed of labeling performance improvement (c.f.~\eqref{eq:reward}~\eqref{eq:value}), and $\gamma$ controls how greedy the policy would be. When $\gamma=0$, the optimal policy maximizes a myopic objective as in~\cite{grubb2012speedboost}. We choose $\gamma>0$ so that our policy also considers potential future benefit (i.e., fast improvement in later stages).

\section{Learning anytime scene labeling}\label{sec:learn}
To learn anytime scene labeling, we want to seek a policy $\pi^*$ to maximize the expected value function for any state $s_t$ in a MDP framework. Given a training set $\mathcal{D}\allowbreak =\{I^{(m)}, \allowbreak \hat{\mathbf{Y}}^{(m)}\}_{m=1}^M$ with $M$ images, the learning problem is defined as,
	\begin{equation}
	\pi^*(s_t)=\argmax_{\pi}E_D[V_{\pi}(s_t)]=\argmax_{\pi}\frac{1}{M}\sum_{m=1}^M\sum_{\tau=t}^T\gamma^{\tau-t}R(s_\tau,\pi(s_\tau)|I^m), \; \forall t
	\end{equation}
where $E_D$ is the empirical expectation on the dataset $\mathcal{D}$.
The main challenge in solving this MDP is to explore the parametrized action space $\mathcal{A}$ due to its discrete-continuous nature and high-dimensionality. In this work, we design an action generation strategy that proposes a finite set of effective parameters for the actions. We then use the proposed action pool as our discrete action space and develop a least square policy learning procedure to find a high quality policy. 


\subsection{Action proposal generation}\label{subsec:action}

To cope with the parameterized actions, we discretize the parameter space $\varTheta\cup\mathcal{H}$ by generating a finite set of effective and diversified parameter values. Our discretization uses a greedy learning criterion to generate a sequence of actions with instantiated parameters based on the training set $\mathcal{D}$. 

Specifically, we start from $s_0$ for all the training images, and generate a sequence of actions and states (which corresponds to a sequence of hierarchical models) as follows. At step $t$, we first discretize $\varTheta$ by uniformly sample the 1D space. For the weak learner space $\mathcal{H}$, we generate a set of weak learners by minimizing the following regression loss as in the Greedy Miser method~\cite{xu2012greedy}:
\begin{equation}
\alpha_t, \bh_t=\arg\min_{\alpha, \bh} \sum_{i\in D^t}w_i \|\bp_i-\bq^{t-1}_i-\alpha \bh(\bbf^{t}_{i})\|^2+\lambda (c_{h_t}+c_{f_t})
\end{equation}	
where $D^t=\{i|Z_m^t(i)=1,m=1,\dots,M\}$ is the set of all active nodes at step $t$ in all $M$ images, and $\bp_i$ and $\bq^{t-1}_i$ are the ground-truth marginal and the previous marginal prediction on node $i$ respectively. The second term regularizes the loss with the cost of applying the weak learner and a weight parameter $\lambda$ controls its strength. We obtain several weak learner $\alpha_t\bh_t$ by varying the value of $\lambda$. 
From these discretized actions, denoted by $\mathcal{A}_s^t$, we then select a most effective action using our reward function, $a_t^0=\arg\max_{a_t\in\mathcal{A}_s^t}\allowbreak E_D[R(s_t,a_t|I)]$. We continue this process until step $T$ based on a held-out validation set, and $\{a_t^0\}_{t=1}^T$ is a sampled action sequence. 

To increase the diversity of our discrete action candidates, we also apply the same action proposal generation method to different subsets of images. The image subsets are formed by K-means clustering and we refer the reader to the supplementary for the details. Finally we combine all the generated discrete action sequences as our action candidates to form a new discrete action space $\mathcal{A}^d$, which is used for learning our policy. 

\subsection{Least-square policy iteration for solving MDP}\label{subsec:iter}
In order to find a high-quality policy $\pi_d^*$ on $\mathcal{A}^d$, we adopt an approximate least-square policy iteration approach and learn a parametrized Q-function~\cite{lagoudakis2003least,Weiss2013}, which can be generalized to the test scenario. Specifically, we use a linear function to approximate the Q-function, and the approximate Q and corresponding policy can be written as
\begin{align}
\hat{Q}(s_t,a_t)&=\eta^{T}\phi(s_t,a_t), \\ \pi_d(s_t)&=\arg\max_{a_t\in\mathcal{A}^d}\hat{Q}(s_t,a_t)
\end{align} 
where $\phi(s_t,a_t)$ is the meta-feature of the model computed from the current state $s_t$ and action $a_t$. $\eta$ is the linear coefficient to be learned. We will discuss the details of our meta-feature in Sec~\ref{subsec:detail}.

Our least-square policy iteration procedure includes
the following three steps, which starts from an initial policy $\pi_d^0$ and iteratively improves the policy.

\textbf{A. Policy initialization} We initialize the policy $\pi_d^0$ by a
greedy action selection that optimizes the average immediate reward on the training set at each time step. Specifically, at each t, we choose $\pi_d^0(s_t) = \arg\max_{a_t\in \mathcal{A}^d} E_D[R(s_t,a_t |I )]$. 


\textbf{B. Policy evaluation.} 
Given a policy $\pi_d^n$ at iteration $n$, we execute the policy for each training example to generate a trajectory $\{(s_t^m,a_t^m)\}_{m=1}^M$. 
We then compute the value function of the policy recursively based on $Q_{\pi}(s_t,a_t)=R(s_t,a_t|I)+\gamma Q_{\pi}(s_{t+1},a_{t+1})$. 
As in~\cite{Weiss2013}, we only consider the non-negative contribution of $Q_\pi$, which allows early stop if the reward is no longer positive,
\begin{equation}
Q_{\pi}(s^m_t,a^m_t)=R(s^m_t,a^m_t)+\gamma [Q_{\pi}(s^m_{t+1},a^m_{t+1})]_{+}
\end{equation}

\textbf{C. Policy improvement.} Given a set of trajectories $\{(s_t^m,a_t^m)\}_{m=1}^M$ and the corresponding Q-function value $\{Q_\pi(s_t^m,a_t^m)\}_{m=1}^M$, we update the linear approximate $\hat{Q}$ by solving the following least-square regression problem:
\begin{equation}\min_{\eta} \beta\|\eta\|^2+\frac{1}{TM}\sum_m\sum_{t}\Big(\eta^T\phi(s_t^m,a_t^m)-Q_{\pi}(s_t^m,a_t^m)\Big)^2
\end{equation}
where the iteration index $n$ is omitted here for clarity. Denote the solution as $\eta^*$, we can compute the new updated policy $\pi_d^{n+1}(s_t)=\arg\max_{a_t}\eta^*\phi(s_t,a_t)$. We also add a small amount of uniformly distributed random noise to the updated policy as in~\cite{Karayev2014}. 
We perform policy evaluation (Step B) and improvement iteration (Step C) several times until the segmentation performance does not change in a held-out validation set. 

During the test, we apply the learned policy $\pi_d$ to an test image, which produces a trajectory $\{(s_0,a_0),\allowbreak{} (s_1,a_1),\allowbreak{} \dots,(s_T,a_T)\}$. The state sequence defines a coarse-to-fine scene labeling process based on the generated hierarchical models. For any given cost budget, we can stop the scene labeling process and use the leaf-node marginal label distributions (i.e., taking the most likely label) to make a pixel-wise label prediction for the entire image. More detailed discussion of the test-time procedure can be found in the supplementary. 




\section{Experiments}
We evaluate our method on three publicly available datasets, including CamVid~\cite{BrostowSFC:ECCV08}, Standford Background~\cite{Gould:ICCV09} and Siftflow~\cite{liu2011nonparametric}. We focus on CamVid~\cite{BrostowSFC:ECCV08} as it provides more complex scenes with multiple foreground object classes of various scales. We refer the reader to supplementary material for the details of datasets.
\subsection{Implementation details}~\label{subsec:detail}
\textbf{Feature set and action proposal:} We extract 9 different visual features (Semantics using Darwin~\cite{JMLR2012}, Geometric, Color and Position, Texture, LBP, HoG, SIFT and hyper-column feature~\cite{hariharan2014hypercolumns,long2014fully}). In action proposal, the weak learner $\alpha_t\bh_t$ is learned as in Sec~\ref{subsec:action} using~\cite{xu2012greedy}. To propose multiple weak learners with a variety of costs, 
we also learn $p$ weak learners sequentially where $p$ is set to 5,10 and 20 empirically and use them as action candidates. 
As for split action, we discretize $\varTheta$ into $\{0,0.3,0.6,1\}$ and we generate a 8-layer hierarchy using~\cite{felzenszwalb2004efficient} as in~\cite{Grubb2013}. In our experiment, we use grid search method and choose the set of hyper-parameters that gives us the optimal pixel-level prediction. 

The cost of each feature type measures the computation time for an entire image. We note that this cost can be further reduced by efficient implementation of local features. The segmentation time is taken into account as an initial cost during the evaluation in order to have a fair comparison with existing methods. More details on cost computation can be found in the supplementary materials. 
\textbf{Policy learning features:} We design three sets of features for $\phi(s_t,a_t)$. The first are computed from marginal distributions on all regions, consisting of the average entropy, the average entropy gap between previous marginal estimation and current marginals, two binary vectors of length 9 to indicate which feature set has been used and which unseen feature set will be extracted respectively, and one vector for the statistics of difference in current marginal probabilities of the top two predictions. The second are region features on active leaf nodes, including the normalized area of active regions in current image, the average entropy of active regions and the average entropy gap between previous and current prediction in active regions. The third layer features consist of the distribution of all regions in hierarchies and the distribution of active regions in hierarchies. 
More details on policy learning features can be found in the supplementary material.

\subsection{Baseline methods}
We compare our approach to two types of baselines as below. We also report the state-of-the-art performances on three datasets.\\
$\bullet$ Non-anytime CRF-based methods using the full feature set: 
1) A fully-connected CRF (DCRF) model~\cite{Koltun2011} whose data term is learned on finest layer of segmentation trees; 
2) A Hierarchical Inference Machine (HIM) implemented by following the algorithm in~\cite{munoz2010stacked}; 3) A pixel-level dense CRF model with superpixel higher-order terms (H-DCRF) as in~\cite{vineet2014filter}. They prove to be strong baselines for scene labeling tasks. \\ 
$\bullet$ Three strong anytime baselines, including a Static-Myopic (S-M), a Random Selection (RS) and a static-myopic feature selection (F-SM) anytime model. The static-myopic method (S-M) 
learns a fixed sequence of actions by maximizing immediate rewards on training set (cf. $\{a_t^0\}_{t=1}^T$ in Sec~\ref{subsec:action}). 
The random selection method (RS) uses our action pool and randomly takes an action at each step. 
The feature selection method (F-SM) uses the DCRF above as its model and greedily selects features that maximize the immediate rewards. 
We note that the baselines utilize some state-of-the-art feature selection methods such as~\cite{xu2012greedy,grubb2012speedboost}, and our RS baseline is built on the learned high-quality action pool. 

\subsection{Results}

We report the results of our experiments on anytime scene labeling in three parts: 1) overall comparison with the baselines and the state-of-the-art methods on CamVid. 2) detailed analysis of anytime property on CamVid. 3) results on Stanford Background and Siftflow datasets.

\textbf{Overall performance on CamVid.}
We first show the quantitative results of our method and compare with state-of-the-art methods in Fig~\ref{fig:percentage_figure}$.(a)$ and Table~\ref{tab:result_seman}. We compute the accuracy and Intersection-Over-Union(IOU) score of semantic segmentation on CamVid. Note that here we report the performance of anytime methods at the time budget of $T_{DCRF}$, which is the average prediction time of DCRF.
In Table~\ref{tab:result_seman}, we can see that our method achieves better performance than DCRF in terms of per-pixel accuracy and IOU score, and DCRF is a strong baseline since it uses the full feature set. Our per-pixel accuracy is comparable to the HIM, which uses the most complex model and full feature set, while we achieve similar performance with about $50\%$ of its computation cost (See below for details).
In addition, we outperform all the rest of state-of-the-art methods~\cite{TigheL13,Grubb2013}, especially in terms of average per-class accuracy ($5.4\%$ to $9\%$ absolute gap). Moreover, we achieve similar or slightly better performance w.r.t. the methods that use additional information such as Structure-from-Motion(SfM) of video sequence~\cite{BrostowSFC:ECCV08,sturgess2009combining} or pre-trained object detectors~\cite{floros2011multi}.

\begin{figure}[t]
	\centering
	\centering
	{
		\begin{tabular}{ccc}
			\hspace{-0.mm}\includegraphics[width=0.32\textwidth]{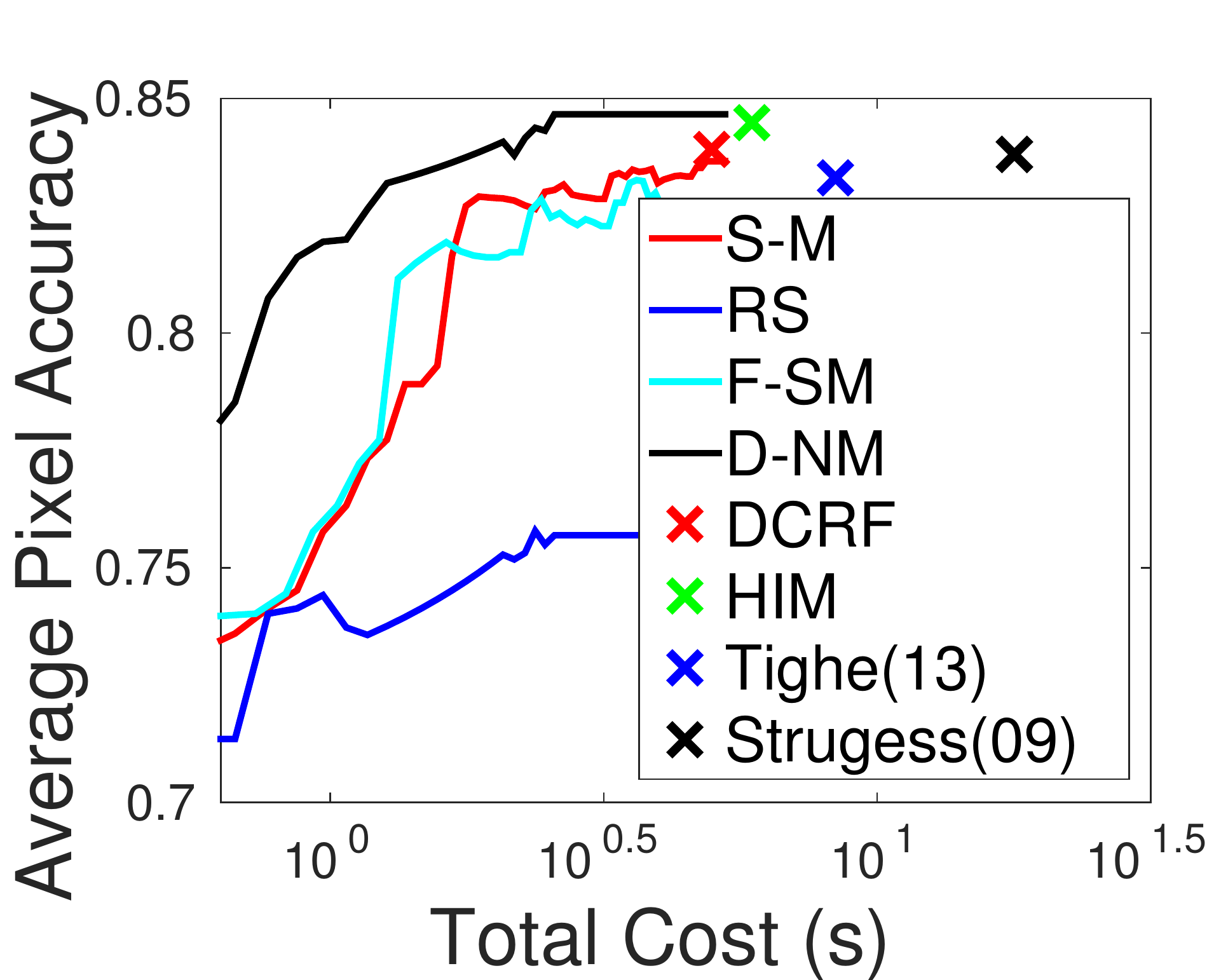} &
			\hspace{-0.mm}\includegraphics[width=0.30\textwidth]{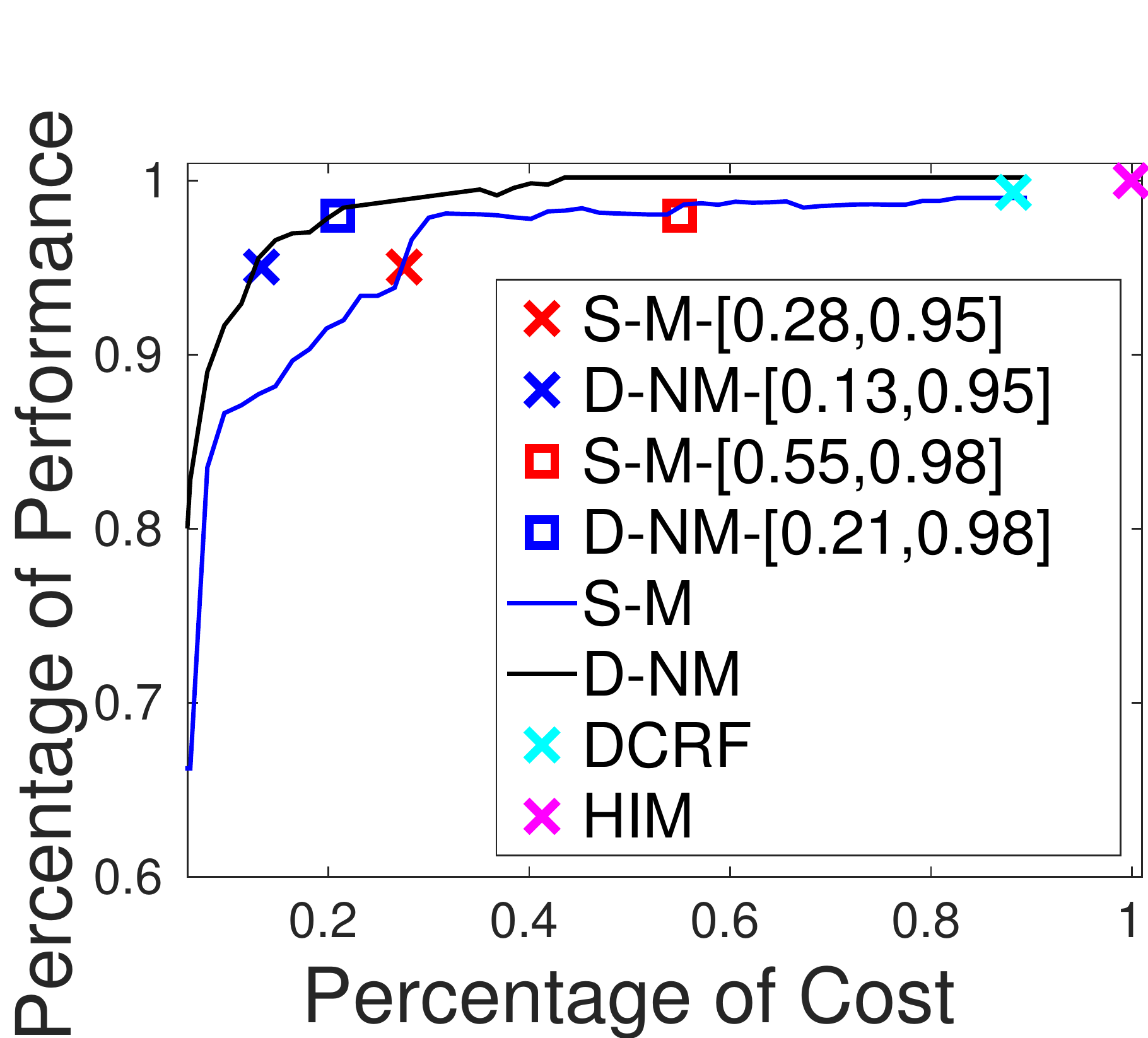} &
			\hspace{-0.mm}\includegraphics[width=0.33\textwidth]{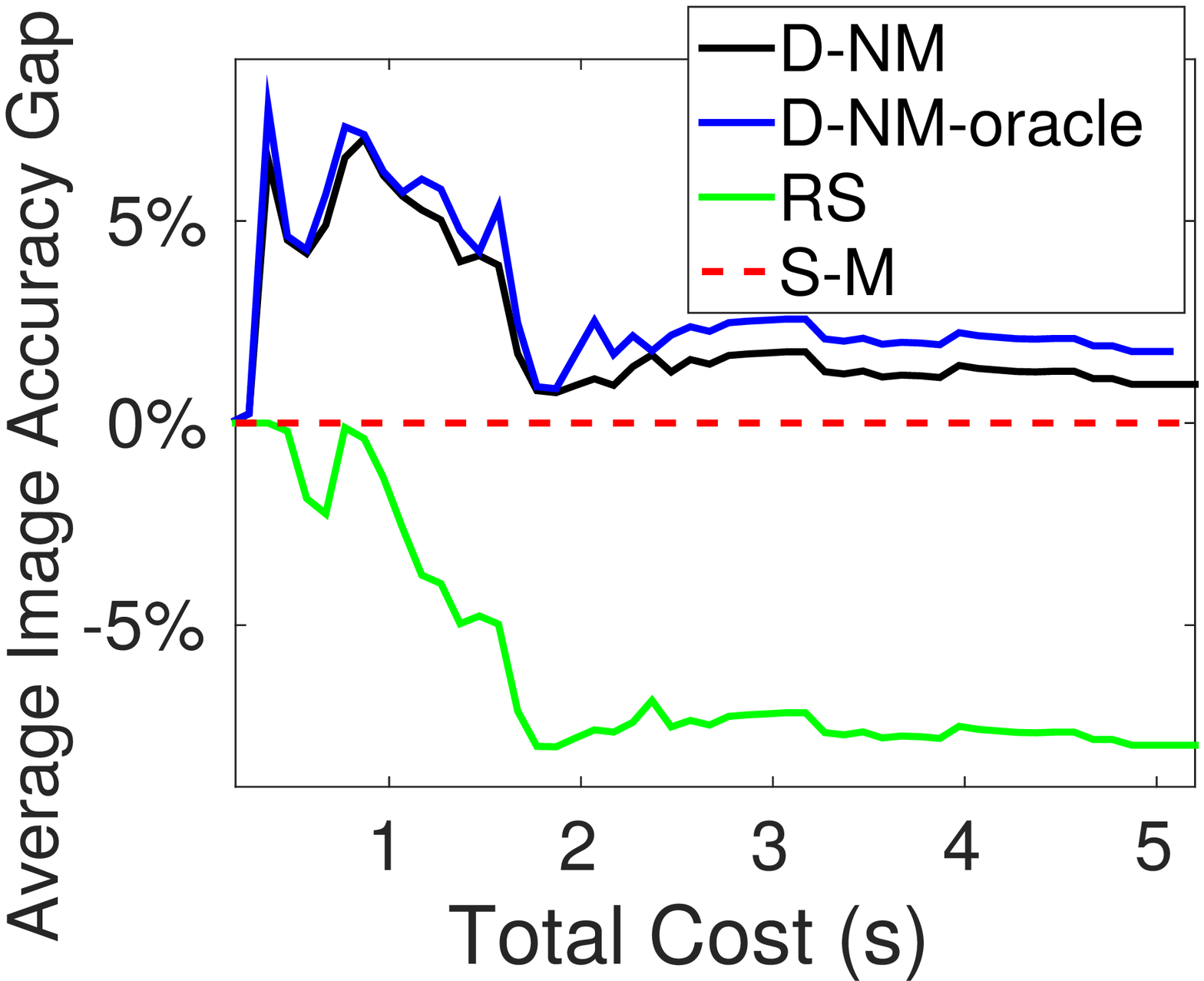} \\ \vspace{-4mm}
			\small{$(a)$} & \small{$(b)$} & \small{$(c)$} \\ 
		\end{tabular}
	}
	\caption{\small
		{{$(a):$ Average pixel accuracy v.s. cost; $(b):$ percentage of performance v.s. percentage of cost; $(c):$ average per-image accuracy gap v.s. total cost in CamVid. Our D-NM consistently outperforms S-M in all figures and achieves full performance using about $50\%$ total cost. Moreover, it outperforms all anytime baselines consistently and achieves better performance w.r.t. non-anytime state-of-the-art.}
		}}\label{fig:percentage_figure}
\end{figure}

\begin{table}[t]
	\centering 
	{\small
		\begin{tabular}{c|c|c|c|c|c|c|c|c|c}
			\multirow{2}{*}{\textbf{CamVid}} & \multirow{2}{*}{Tighe~\cite{TigheL13}} & \multirow{2}{*}{SIM~\cite{Grubb2013}} & Video & Detector & Video & \multirow{2}{*}{DCRF} & H- & \multirow{2}{*}{HIM} & D-NM \\
			& & & ~\cite{BrostowSFC:ECCV08}& ~\cite{floros2011multi} & ~\cite{sturgess2009combining} & & DCRF & &(ours) \\ \hline
			\textbf{Pixel} & 83.3 & 81.5 & 69.1 & 83.2 & 83.8 & 83.2 & 83.9 & 84.5 & \bf 84.7 \\ \hline
			\textbf{Class} & 51.2 & 54.8 & 53.0 & 59.6 & 59.2 & 59.8 & 60.0 & \bf 60.5 & 60.2 \\ \hline	
			\textbf{IOU} & NA & NA & NA & \bf 49.3 & 49.2 & 46.3 & 48.4 & \bf 49.3 & 48.8 \\ \hline
		\end{tabular}
	}
	\caption
	{
		\small{Performance comparison on CamVid. D-NM outperforms~\cite{TigheL13,Grubb2013}, especially in average class accuracy. Our results are comparable to~\cite{BrostowSFC:ECCV08,sturgess2009combining,floros2011multi} that use additional information. We achieve a performance similar to HIM and DCRF with less cost.}
	}\label{tab:result_seman}
\end{table}


We conduct comparisons on the anytime performance of our methods and baselines in Fig~\ref{fig:percentage_figure}$.(a)$ and $(b)$. We introduce a plot $(b)$ showing all the performance and cost values w.r.t the HIM and its prediction cost since it is the state-of-the-art and most costly. Specifically, Figure~\ref{fig:percentage_figure}$.(b)$ shows the percentage of average pixel-level accuracy v.s. percentage of total cost curves of our methods and baselines w.r.t the HIM. We note that this illustration is invariant to the specific values of prediction cost/time, and shows how the accuracy improves with increasing cost.


We first show comparison of our method with all the anytime and non-anytime baselines in Figure~\ref{fig:percentage_figure}$.(b)$, which also highlights two sets of intermediate results. Our dynamic policy D-NM achieves the $90\%$ of performance using only around $10\%$ cost and outperforms the S-M consistently. Specifically, D-NM achieves similar performance with around half S-M test-time cost ($13\%$ and $21\%$ v.s. $28\%$ and $55\%$). 
Moreover, D-NM achieves the full performance of HIM with around $50\%$ total cost while S-M saturates at a lower accuracy. 
We refer the reader to the supplementary material for examples of our anytime output with specific actions.

\begin{figure}[t!]
	\centering
	\centering
	{
		\begin{tabular}{cccc}			
			\hspace{-0.5mm}\includegraphics[width=0.25\textwidth]{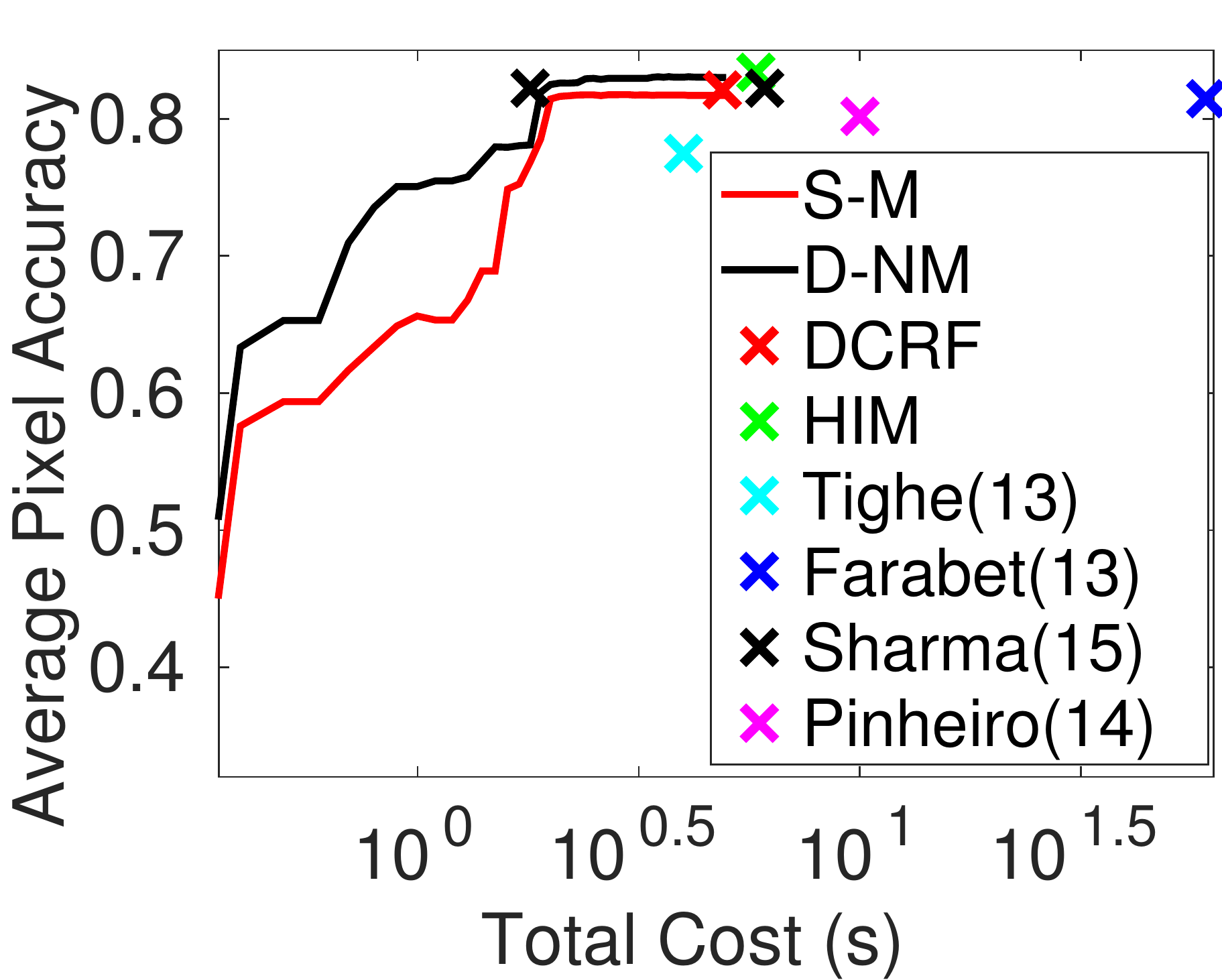} &\hspace{-0.5mm}\includegraphics[width=0.25\textwidth]{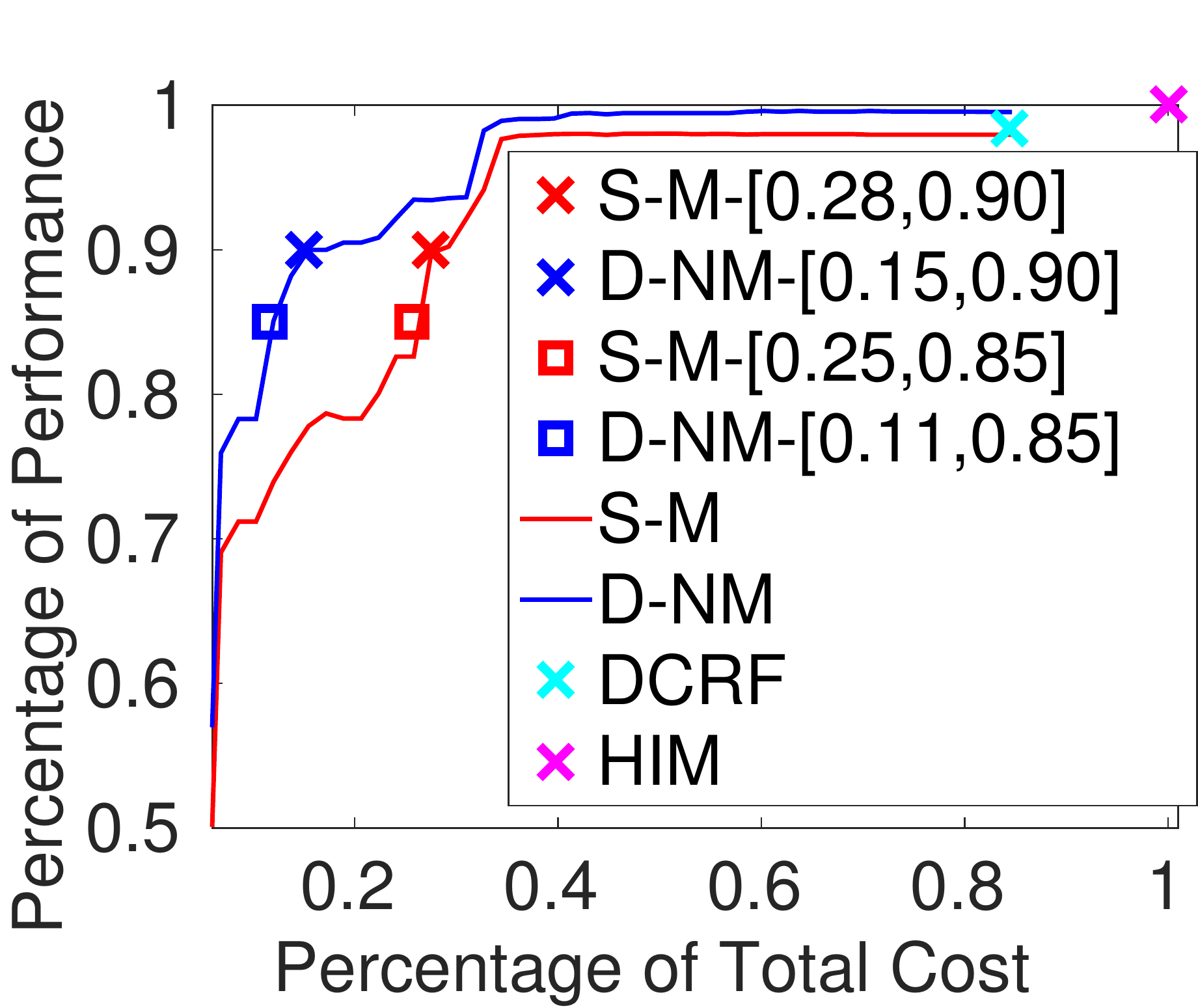}
			&\hspace{-0.5mm}\includegraphics[width=0.25\textwidth]{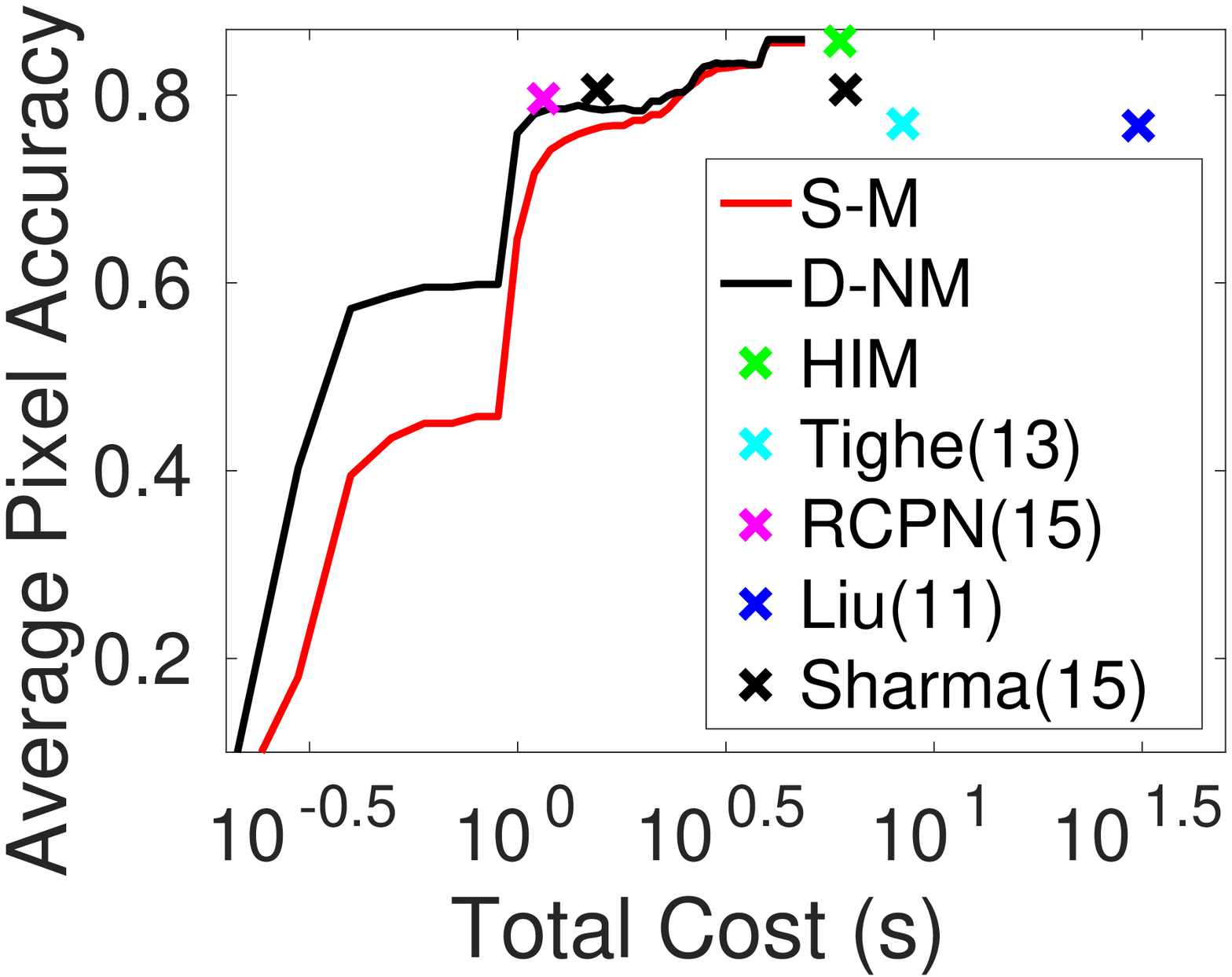} &\hspace{-0.5mm}\includegraphics[width=0.25\textwidth]{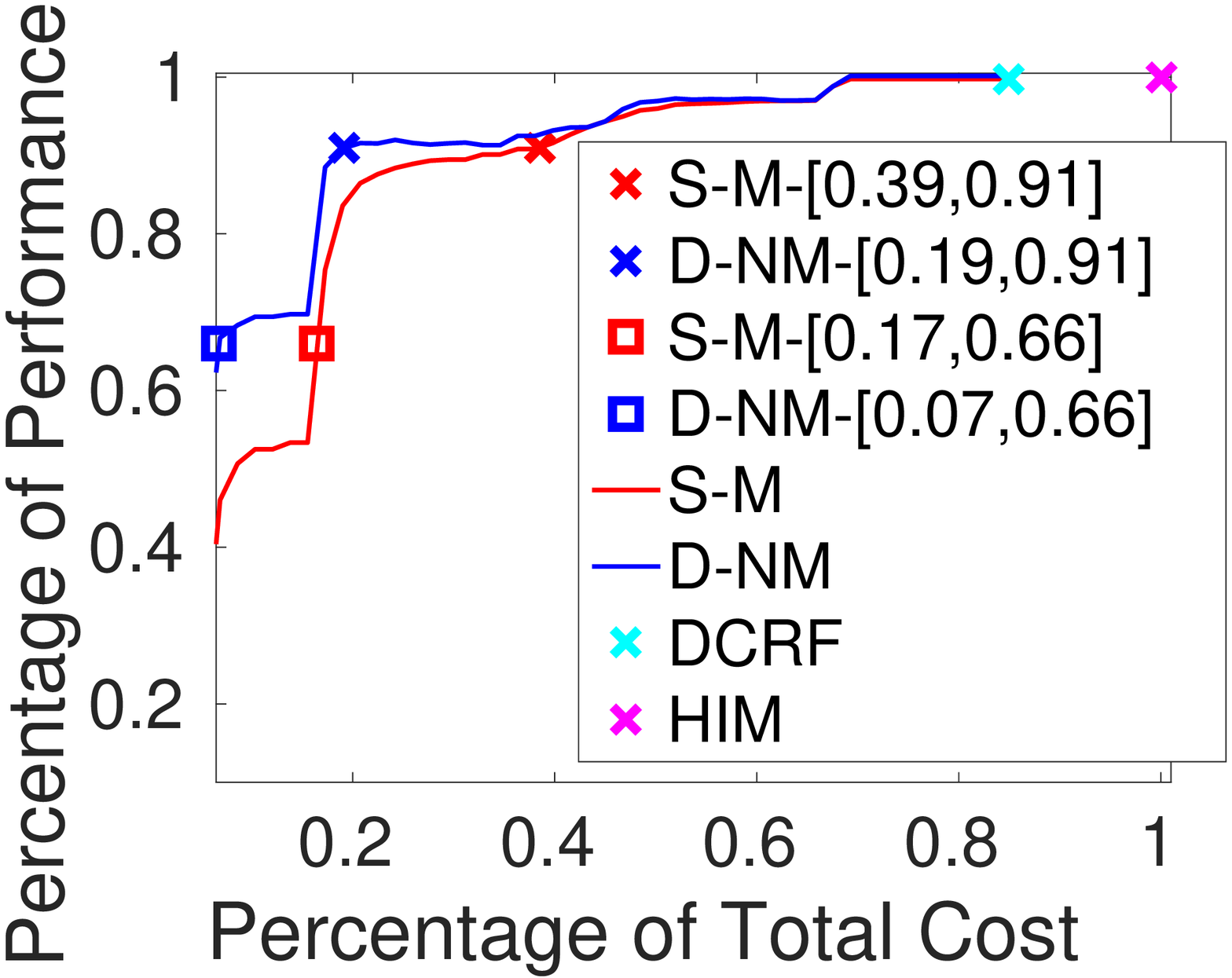}\\
			$(a)$ & $(b)$ &$(c)$ &$(d)$ 
		\end{tabular}
	}
	\caption{\small
		{{ Average pixel accuracy as a function of cost and the percentage performance v.s. percentage of cost in SBG (a,b) and Siftflow (c,d), respectively. D-NM achieves similar performance with less cost. Cost of related work is from~\cite{sharma2015deep}.}
		}}\label{fig:percentage_figure_sbgsiftflow}
	\end{figure}


\textbf{Anytime property analysis on CamVid.}
We analyze the anytime property of our method by comparing to three different baselines. First, we validate the importance of encoding model complexity in anytime prediction model by comparing with F-SM (fixed model with feature selection). 
Second, we evaluate the effectiveness of policy learning by comparing with RS (random search on the same action space). Results of these two comparisons can be viewed in Figure~\ref{fig:percentage_figure}$.(a)$. Finally, we explore the effectiveness of action space exploration by generating the oracle results of D-NM on CamVid test set in Figure~\ref{fig:percentage_figure}$.(c)$.


Figure~\ref{fig:percentage_figure}$.(a)$ shows that the D-NM outperforms all baselines consistently and generates superior results under the same cost. RS is almost always the worst and far below D-NM, which shows our policy learning is important and effective to achieve better trade-offs between accuracy and cost. F-SM is slightly above S-M at the beginning and always below D-NM. Moreover, due to the limited representation power of its fixed model, F-SM quickly stabilizes at a lower performance. This demonstrates the benefits of joint feature and model selection in our method. 
We also visualize results of other methods {(crossings)} and show that we can achieve better performance more efficiently. These results evidence that our method can learn a better representation for anytime scene parsing. Detailed averaged IOU score and labeling loss as a function of cost, area-under-average-accuracy table can be viewed in supplementary materials.

Figure~\ref{fig:percentage_figure}$.(c)$ shows the average per-image accuracy gap w.r.t the S-M method as a function of total cost. We note that D-NM always achieves superior performance to the S-M. We also visualize the oracle performance of D-NM. D-NM-oracle is always above S-M, which proves the effectiveness of action space exploration. Also, the early stop of oracles shows that more features or complex models will not introduce further segmentation improvement. Our D-NM is only slightly below D-NM-oracle, which shows the effectiveness of policy learning. 

\begin{table}[t]
	\footnotesize
	\centerline 
	{
		\begin{tabular}{c|c|c|c|c|c|c|c|c|c}
			\multirow{2}{*}{\textbf{SBG}} & \multirow{2}{*}{RCPN~\cite{sharma2014recursive}} & Tighe & Gould & Farabet & Pinheiro & Sharma & H- & \multirow{2}{*}{S-M} & D-NM\\ 
			& & ~\cite{TigheL13} & ~\cite{Gould:ICCV09}  & ~\cite{farabet2013learning} & ~\cite{pinheiro2014recurrent} & ~\cite{sharma2015deep} & DCRF & &(ours) \\ \hline
			\textbf{Pixel} & 81.8 & 77.5 & 76.4 & 81.4 & 80.2 & 82.3 & 82.6 & 81.7 & \bf 83.0 \\ \hline
			\textbf{IOU} & 61.3 & NA & NA & NA & NA & 64.5 & \bf64.7 & 61.4 & \bf 64.7 \\ \hline
		\end{tabular}
	}
	\caption
	{
		\small{Semantic segmentation results on Stanford background dataset. We can achieve better performance w.r.t state-of-the-art methods.}
	}\label{tab:result_seman_sbg}
\end{table}

\begin{table}[t]
	\footnotesize
	\centerline 
	{
		\begin{tabular}{c|c|c|c|c|c|c|c|c|c|c|c}
			\multirow{2}{*}{\textbf{Siftflow}} & RCPN & Yang & Pinheiro & Liu & Tighe & FCN & Farabet & Sharma & H- & \multirow{2}{*}{S-M} & D-NM \\ 
			& ~\cite{sharma2014recursive} & ~\cite{yang2014context} & ~\cite{pinheiro2014recurrent} & ~\cite{liu2011nonparametric} & ~\cite{TigheL13} & ~\cite{long2014fully} & ~\cite{farabet2013learning} & ~\cite{sharma2015deep} & DCRF & &(ours) \\ \hline
			\textbf{Pixel} & 79.6 & 79.8 & 77.7 & 76.7 & 77.0 & 85.7 & 78.5 & 80.8 & \bf 85.8 & \bf 85.8 & \bf 85.8 \\ \hline
			\textbf{IOU} & 26.9 & NA & NA & NA & NA & \bf 36.7 & NA & 30.7 & \bf 36.7 & 35.8 & \bf 36.7 \\ \hline
		\end{tabular}
	}
	\caption
	{\small{Semantic segmentation results on Siftflow dataset. We can achieve comparable/better performance w.r.t. state-of-the-art methods.}
	}\label{tab:result_seman_siftflow}
\end{table}

\textbf{Stanford Background.}
Results on Stanford Background dataset~\cite{Gould:ICCV09} are shown in Table~\ref{tab:result_seman_sbg}. D-NM outperforms existing work in terms of pixel-level accuracy and IOU score. We visualize the anytime property in Figure~\ref{fig:percentage_figure_sbgsiftflow}.$(a)$ and $(b)$. Figure~\ref{fig:percentage_figure_sbgsiftflow}$.(a)$ shows that D-NM achieves the state-of-the-art performance {(crossings)} more efficiently while S-M stops at a lower performance. Figure~\ref{fig:percentage_figure_sbgsiftflow}$.(b)$ {highlights two sets of intermediate results} and shows that D-NM generates similar results with about half of the S-M cost ($11\%$ and $15\%$ v.s. $25\%$ and $28\%$). 






\textbf{Siftflow.}
We report our results on Siftflow dataset~\cite{liu2011nonparametric} in Table~\ref{tab:result_seman_siftflow}. Again, D-NM achieves the state-of-the-art in terms of pixel level accuracy and IOU score. Figure~\ref{fig:percentage_figure_sbgsiftflow}$.(c)$ shows its anytime performance curves and Figure~\ref{fig:percentage_figure_sbgsiftflow}$.(d)$ also highlights two sets of intermediate results. 
We can see that D-NM achieves the state-of-the-art performance {(crossings)} more efficiently, and produces similar accuracy with much less cost. 

\section{Conclusion}
In this paper, we presented a dynamic hierarchical model for anytime semantic scene segmentation. Our anytime representation is built on a coarse-to-fine segmentation tree, which enables us to select both discriminative features and effective model structure for cost-sensitive scene labeling. 
We developed an MDP formulation and an approximated policy iteration method with an action proposal mechanism for learning the anytime representation.
The results of applying our method to three semantic segmentation datasets show that our algorithm consistently outperforms the baseline approaches and the state-of-the-arts.
This suggests that our learned dynamic non-myopic policy generates a more effective representation for anytime scene labeling. 
\clearpage

\bibliographystyle{splncs}
\bibliography{egbib}
\end{document}